\documentclass[runningheads]{llncs}
\usepackage[T1]{fontenc}
\usepackage{graphicx}
\usepackage{amsfonts}
\usepackage{tabularx}
\usepackage{bm}
\usepackage[linesnumbered,lined,boxed,commentsnumbered, ruled,vlined]{algorithm2e}
\usepackage{xcolor}

\SetCommentSty{mycommfont}
\begin{document}
\title{Using Part-based Representations for Explainable Deep Reinforcement Learning}
%
%
\author{Manos Kirtas\orcidID{0000-0002-8670-0248} \and
Konstantinos Tsampazis \and
Loukia Avramelou \and
Nikolaos Passalis\orcidID{0000-0003-1177-9139} \and
Anastasios Tefas\orcidID{0000-0003-1288-3667}}
\authorrunning{M. Kirtas, K. Tsampazis, et al.}
%
\institute{Computational Intelligence and Deep Learning Research Group \\
School of Informatics, Aristotle University of Thessaloniki, Greece.\\
\email{\{eakirtas, tsampaka, avramell, passalis, tefas\}@csd.auth.gr}}
\maketitle              
\begin{abstract}
Utilizing deep learning models to learn part-based representations holds significant potential for interpretable-by-design approaches, as these models incorporate latent causes obtained from feature representations through simple addition. However, training a part-based learning model presents challenges, particularly in enforcing non-negative constraints on the model's parameters, which can result in training difficulties such as instability and convergence issues. Moreover, applying such approaches in Deep Reinforcement Learning (RL) is even more demanding due to the inherent instabilities that impact many optimization methods. In this paper, we propose a non-negative training approach for actor models in RL, enabling the extraction of part-based representations that enhance interpretability while adhering to non-negative constraints. To this end, we employ a non-negative initialization technique, as well as a modified sign-preserving training method, which can ensure better gradient flow compared to existing approaches. We demonstrate the effectiveness of the proposed approach using the well-known Cartpole benchmark.
\end{abstract}

\section{Introduction}
\label{sec:intro}

Deep Reinforcement Learning (RL) has achieved state-of-the-art performance in various applications, including robotics~\cite{kober2013reinforcement,9351818}. However, the use of RL agents in critical environments, where safety is highly prioritized, is hindered due to the limited transparency of the models. Extracting the rationale of a deep learning (DL) model in a human-interpretable way remains a challenging task, but doing so would be highly useful for improving both the performance and trustworthiness of the model, as well as preventing failures~\cite{10.1145}. To this end, \textit{post-hoc} explanation methods have been extensively studied over the years, providing rationales for the predictions of the model~\cite{10.1007,bach2015pixel}. However, such approaches cannot always provide a reliable explanation~\cite{10.11488,Zhou_2022}, with \textit{pre-hoc} methods for explainable AI gaining increasing attention recently~\cite{10.1214}. The \textit{pre-hoc} approaches aim to design inherently explainable models, providing a transparent mechanism to the decision-making process in such a way that one can calibrate user trust and predict the system's capabilities.            

To this end, extracting a part-based representation of deep learning models provides great potential for interpretable-by-design approaches, since they are based on the simple addition of latent causes acquired from feature representations, making models easily interpretable by human actors due to the elimination of canceling neurons~\cite{LEMME2012194,7310882}. However, training a part-based learning model is challenging since it requires non-negative constraints to the model's parameters, leading to training difficulties, such as instabilities and convergence issues~\cite{6783731}. Furthermore, existing approaches for part-based learning are limited, e.g., applied solely on autoencoders~\cite{7310882,8051252}, and models that are not usually used in DL models, such as Pyramid Neural Networks~\cite{8489216,7966057}, resulting in a significant performance degradation~\cite{6783731}, making them unsuitable for RL.

In this work, we propose a non-negative training approach for actor models in RL approaches, allowing for extracting part-based representations that can provide increased interpretability, while also building upon non-negative constraints that are known to be conceptually tied to human cognition~\cite{lee1999learning,10.1093}.  To this end, the proposed method employs a non-negative initialization method, along with an appropriately modified sign-preserving training method.  More specifically, we propose using an exponential distribution-based non-negative initialization method for the actor model. Then, we introduce a sign-preserving alternative of Stochastic Gradient Ascent (SGA) that is used to train the actor model in a non-negative manner. The proposed optimization method allows better gradient flow, compared to existing clipping-based approaches, reducing the phenomenon of vanishing gradients and increasing the stability of the training process. As a result, the proposed pipeline enables more efficient training of inherently explainable models that are based on the non-negative part-based representation of the actor. Note that even though the proposed method is presented within the Proximal Policy Optimization (PPO)~\cite{schulman2017proximal} algorithm, this is without loss of generality and could be readily adapted to any other Deep RL approach. We demonstrate the effectiveness of the proposed method in a traditionally used benchmark, named Cartpole, in a high-fidelity 3D robotics simulation.

The remainder of this paper is structured as follows. The proposed method is introduced and described in detail in Section~\ref{sec:proposal}, while the experimental evaluation is provided in Section~\ref{sec:results}. Finally, conclusions are drawn in Section~\ref{sec:conclusions}.

\section{Proposed Method}
\label{sec:proposal}

In this work, we focus on training non-negative agents using policy gradient-based approaches, such as the PPO algorithm~\cite{schulman2017proximal}, but without loss of generality, since the proposed method can also be directly applied to other RL methods as well, such as Q-learning based approaches. More specifically, PPO utilizes actor-critic networks, where the actor model decides which action should be taken, with  its parameters denoted as \(\bm{\theta}\). On the other hand, the critic network, equipped with parameters \(\bm{\tilde{\theta}}\), informs the actor about the quality of its actions and guides the actor on how to adjust them during training. The PPO method trains the actor based on the policy gradient approach, while the critic evaluates the actions by computing the corresponding state/action values. For simplicity, we assume that both models have the same number of layers.

First, we propose an initialization scheme for the parameters of the actor model using an exponential distribution to ensure a positive-only initialization. Then, to train the actor model in a part-based representation manner, we propose a non-negative optimization approach based on Stochastic Gradient Ascent (SGA), ensuring that the canceling neurons of the network will be diminished by constraining parameters to the non-negative space, making them more easily interpretable by humans. Thus, only the actor network is trained in a non-negative manner, since it is responsible for the actions made by the agent during the deployment and, as a result, it is the one that needs to be explainable during deployment. Constraining to positive values only the parameters of the actor model, without similarly restricting the critic model, allows for reducing the risk of convergence issues that usually arise in non-negative neural networks~\cite{6783731}. At the same time, this does not reduce the interpretability of the actor model during deployment, since the critic is only used during the training process.

In the case of RL approaches with a policy gradient, the actor model is trained to learn a policy, \(\pi(a|\mathbf{s})\), by observing the state \(s\) and returning the probability of selecting the action \(a\). The groundbreaking results of the PPO are attributed to the constraints utilized in the policy parameter steps. More precisely, PPO employs the action probability ratio between the policy parameterization formulated as:
\begin{equation}
    r_t(\bm{\theta}) = \frac{\pi_{\bm{\theta}}(a|\mathbf{s}_t)}{\pi_{\bm{\theta}_{old}}(a|\bm{s}_t)} \in \mathbb{R},
\end{equation}
where \(\pi_{\bm{\theta}}(a|\bm{s}_t)\) is the probability that policy \(\pi\), with actor's parameters \(\bm{\theta}\), selecting an action \(a\) when the agent observes environment state \(\bm{s}_t\) at time step \(t\) and the previous step parameters are denotes as \(\bm{\theta}_{old}\). Using the clipped version of the ratio \(r_{t}(\bm{\theta})\) around the value of 1 within \(\epsilon\), the policy exploration can be constrained to the close vicinity of the parameter space. The clipped policy ratio is defined as:
\begin{equation}
    r_t^{clip}(\bm{\theta}) = \textrm{clip}(r_t(\bm{\theta}), 1-\epsilon, 1+\epsilon) \in [1-\epsilon, 1+\epsilon],
 \end{equation}
where \(\epsilon\) is the constraint range of policy update and by default is set to \(\epsilon=0.2\), while the clip function is defined as:
\begin{equation}
    \textrm{clip}(x, m, M) = \max(\min(x, M), m) \in [m, M].
\end{equation}
The final objective function of the PPO is defined as:
\begin{equation}
    L^{actor}(\mathbf{s}_t; \bm{\theta}, \bm{\tilde{\theta}}) = \mathbb{E}_t\left[ \min\left(r_t^{clip}(\bm{\theta})A_t(\bm{\tilde{\theta}}), r_t^{clip}(\bm{\theta})A_t(\bm{\tilde{\theta}})\right) \right] \in \mathbb{R},
\end{equation}
where \(A_t(\bm{\tilde{\theta}})\) is the advantage. In this work, we use the General Advantage Estimation (GAE)~\cite{schulman2017trust} approach.
The Temporal Difference (TD) residual for each time step \(t\) is calculated as:
\begin{equation}
    \delta_t(\bm{\tilde{\theta}}) = R_t + \gamma V_{\bm{\tilde{\theta}}_t}^{\pi}(\mathbf{s}_{t+1}) - V_{\bm{\tilde{\theta}}_t}^{\pi}(\mathbf{s}_t) \in \mathbb{R},
\end{equation}
where \(R_t\) is the reward the agent receives at time step \(t\), \(V_{\bm{\tilde{\theta}}_t}^{\pi}(\mathbf{s}_t) \) is the value estimation predicted by the critic policy \(\pi\) for current state \(s_t\) based on critic parameter \(\bm{\tilde{\theta}}_t\), \(\gamma\) is the discount factor and \(\lambda\) is the smoothing parameter. In this work, we use  \(\gamma=0.99\) and \(\lambda=0.95\). Then, the advantage \(A_t\) is defined as:
\begin{equation}
    A_{t}(\bm{\tilde{\theta}}) = \sum_{i=0}^{n-t}\gamma^{i}\lambda^{i}\delta_{t+i}(\bm{\tilde{\theta}}) \in \mathbb{R},
\end{equation}
where \(n\) is the total number of steps within an episode and \(t\) is the time step.

On the other hand, the critic network is typically trained to minimize the temporal difference between the returns and it is formulated as:
\begin{equation}
    L^{critic} = \mathbb{E}_t[\delta_t(\bm{\tilde{\theta}})^{2}] \in \mathbb{R}
\end{equation}

Traditionally used initialization schemes, such as Kaiming~\cite{He_2015_ICCV} and Xavier~\cite{glorot10a}, oriented to ANNs that apply the ReLU activation function, initialize the parameters of the \(k\)-th layer around zero, drawing values from a Gaussian distribution, \(\theta\sim\mathcal{N}(0, \sigma_{k})\), where $\theta$ denotes a parameter from $\bm{\theta}$ and the standard deviation depends on the number of inputs and neurons. To this end, even by applying optimization methods that constrain parameters to positive values, it results in a very low variance of parameters during the first epochs of training that can lead to convergence difficulties or even halt the training process. To this end, inspired by~\cite{8051252}, we propose to initialize actor parameters \(\bm{\theta}\) in the positive domain using an exponential distribution given by:
\begin{equation}
   {\theta} \sim \textrm{Exp}(\lambda) =\frac{\ln(U(0, 1))}{\lambda}\ \in \mathbb{R}_{+},
\end{equation}
where \(U(0, 1)\) is a uniform distribution between \((0, 1)\), \(\mathbb{R}_{+}\) denotes the set of positive real values, and \(\lambda\) is the rate parameter of the distribution and it is a hyperparameter that by default is set to \(\lambda=100\). Even though initializing the actor parameters allows us to obtain a part-based representation before training, the traditionally used optimization algorithms, such as SGA, do not allow one to preserve the initial sign of the parameters. Therefore, we propose a sign-preserving optimization method that is based on the SGA. More specifically, we propose a sign-preserving alternative of SGA that modifies the update term, which is attributed to the sign change, ensuring that the trainable parameters will remain non-negative during the training phase. To this end, the actor's parameters are updated as:
\begin{equation}
    \bm{\theta} = \left| \bm{\theta}_{old} + \eta_{a}\frac{\partial L^{actor}}{\partial\bm{\theta}_{old}}\right| , 
\end{equation}
where the \(\eta_{a}\) denotes the learning rate of the actor and \(|\cdot|\) the absolute value operator. 

On the other hand, critic model is trained as usual employing the standard Stochastic Gradient Descent (SGD) denoted as:
\begin{equation}
    \bm{\tilde{\theta}} = \bm{\tilde{\theta}}_{old} - \eta_{c}\frac{\partial L^{critic}}{\partial\bm{\tilde{\theta}}_{old}},
\end{equation}
where the \(\eta_{c}\) defines the learning rate of the critic model and $L^{critic}$ denotes the loss function of the critic. We call the proposed sign preservation method \textit{Absolute Stochastic Gradient Ascent} (ASGA), since it employs the absolute value operator to preserve the positive sign of parameters during the training process. 

The proposed training method is presented algorithmically in Algorithm~\ref{alg:nn_ppo}. More specifically, it receives as input the initialization hyperparameters for the actor model, which is initialized with the exponential distribution using \(\lambda \in \mathbb{R}\), and for the critic model, which is initialized by a normal distribution using \(\bm{\sigma} \in \mathbb{R}^{n}\), where $n$ is the number of layers. The former network is optimized with the proposed sign-preserving alternative of SGA using learning rate \(\eta_{a}\) and the latter using the SGD optimizer with learning rate \(\eta_{c}\). The algorithm outputs both actor's \(\bm{\theta}\) and critic's \(\bm{\tilde{\theta}} \) parameters, ensuring that the actor's parameters will be non-negative. Firstly, the proposed method initializes the parameters for both models (lines 2-3), using an exponential distribution for the actor (line 3), constraining to positive-only parameters, and a normal distribution for critic parameters (line 4). In turn, the method iterates over the environment for \(T_{episodes}\) episodes (lines 6-12), applying the obtained policy (\(\pi_{\bm{\theta}_{old}}\)) for \(T_{steps}\) steps (lines 6-8), collecting the states, trajectories and rewards (line 7). Then, the PPO algorithm calculates the GAE (line 8) and the proposed method trains both actor and critic for \(T_{epochs}\) (line 9-12) using mini-batches (line 10-12). More precisely, the actor's parameters are updated using the proposed sign-preserving gradient ascent (line 11), while the actor is trained as usual by applying the SGD optimizer (line 12). This results in non-negative trained parameters for the actor model, making the part-based representation of the actor feasible.  

\begin{algorithm}
\caption{Non-Negative Actor PPO Training \label{alg:nn_ppo}}    
  \SetKwData{Left}{left}\SetKwData{This}{this}
  \SetKwData{Up}{up}\SetKwFunction{Union}{Union}
  \SetKwFunction{FindCompress}{FindCompress}
  \SetKwInOut{Input}{Input}
  \SetKwInOut{Output}{Output}
  
  \Input{
    \(\lambda \in \mathbb{R}\): the rate parameter of the exponential distribution, \\
    \(\bm{\sigma} \in \mathbb{R}^{n}\): a vector containing the standard deviation for the Gaussian distribution for each layer, \\ 
    \(\eta_{a} \in \mathbb{R}\) and \(\eta_{c} \in \mathbb{R}\): learning rates for actor and critic.\\
  }
  \Output{
     \(\bm{\theta} \) : actor's parameters, and\\
     \(\tilde{\bm{\theta}}\): critic's parameters.\\
  }

  \Begin{
    \For{\(k = 1 \ldots n \)  }{
        \(\theta^{(k)} \sim Exp(\lambda) \) \tcp*{Initialize Actor's Parameters}
        \(\bm{\tilde{\theta}}^{(k)} \sim \mathcal{N}(0, \sigma_k)\) \tcp*{Initialize Critic's Parameters}

    }
    \For{\(i = 1 \ldots T_{episodes}\) }{
        \For{\(j = 1 \ldots T_{iter}\) }{
            Apply policy \(\pi_{\bm{\theta}_{old}}\) and collect state, actions and rewards\;
            \(A_{j}(\bm{\tilde{\theta}}) = \sum_{i=0}^{n-t}\gamma_{i}\lambda_{i}\delta_{t+i}(\bm{\tilde{\theta}})\) \tcp*{Compute Advantage estimates}
        }
        \For{\(j=1 \ldots T_{epochs}\)}{
            \For{every batch }{
                \(\bm{\theta} = \left| \bm{\theta} + \eta_{a}\frac{\partial L^{actor}}{\partial \bm{\theta}}\right|\) \tcp*{Update Actor's Parameters}
                \vspace{1pt}
                \(\bm{\tilde{\theta}} = \bm{\tilde{\theta}} - \eta_{c}\frac{\partial L^{critic}}{\partial\bm{\tilde{\theta}}}\) \tcp*{Update Critic's Parameters}
            }
        }
    }
    \Return \(\bm{\theta}, \bm{\bm{\tilde{\theta}}}\)\;
}
\end{algorithm}

\section{Experimental Evaluation}
\label{sec:results}

We experimentally evaluated the proposed method on the typical RL benchmark Cartpole implemented using the Deepbots framework~\cite{10.1007/978-3-030-49186-4_6,KIRTAS2022579}. More specifically, the simulated environment is composed of a four-wheeled cart that has a long pole attached to it by a free hinge. On the top of the pole, there is a sensor to measure their vertical angle. The pole acts as an inverted pole pendulum and the goal is to keep it vertical by moving the cart forward and backward. 

We applied the PPO algorithm to all evaluated cases, with the actor network getting the observations of the agent as input, consisting of two hidden layers of 10 neurons each, and outputs the action of the agent. Similarly, the critic network gets the observations as input, and it outputs the advantage of each state. The critic network also consists of two hidden layers of 10 neurons. On both networks, we employed the ReLU activation function in the hidden layers, with the actor model setting an upper bound to ReLU at 2 (ReLU2) for better visualization of the model. The observations contain the cart position, velocity on the x-axis, vertical angle of the pole, and pole's velocity at its tip. The available discrete actions at each step are either to move forward or backward. For each step, the agent is rewarded with \(+1\) and each episode ends after \(T_{steps}=195\) steps or earlier if the pole has fallen \(\pm{15}^{\circ}\) off vertical or if the cart has moved more than \(\pm{39}\) centimeters on the x-axis. The networks are optimized for \(T_{episodes} = 10^{4}\) episodes with learning rates equal to \(
\eta_a = 0.1\) and \(\eta_c = 0.003\) for actor and critic, respectively. The PPO iterates for \(T_{iter}=5\) over batches of 8 collected samples.  

We evaluate the proposed method on the aforementioned setup against two baseline methods using: a) Clipping Stochastic Gradient Ascent (CSGA) with Kaiming initialization and b) CSGA with Xavier initialization. These are based on the clipping approach proposed in~\cite{8051252,6783731}, after appropriate adaptation for use in PPO. More specifically, the CSGA applies a clipping function to the typical stochastic gradient ascent and it is formulated as:
\begin{equation}
    \bm{\theta}'= \max \left( 0, \bm{\theta} - \eta\frac{\partial J}{\partial \bm{\theta}} \right),
\end{equation}
where the $\max(\cdot)$ operator is applied element-wise. In all cases, the evaluated optimization methods are used on the actor model, ensuring the non-negativity of its parameters, with the critic model being trained as usual with the SGD optimizer.

\begin{table}[]
    \centering
    \caption{Average and variance of rewards both for training and evaluation phase over 5 runs. \label{tab:results}}
    \begin{tabularx}{\linewidth}{X|XX}
         Method &  Training & Evaluation \\
         \hline
         CSGA (Kaiming Init.) & \(62.83\pm{39.64}\) & \(89\pm{98.59}\)  \\
         CSGA (Xavier Init.) & \(53.67\pm{35.47}\) &  \(58.2\pm{78.4}\)\\
         \textbf{Proposed} & \(\bm{89.45\pm{1.04}}\) & \(\bm{140.4\pm{43.9}}\) \\
    \end{tabularx}
\end{table}

We conducted five evaluation runs for each method using different seeds during the training phase. In Table~\ref{tab:results} the average and variance of the rewards over 5 runs are reported for all the evaluated methods. Both baselines lead to an unstable training process, resulting in high variance reward values at the end of the training. In contrast, the proposed method offers significantly more consistent training, resulting in low-variance rewards after training, as well as higher performance. This behavior is also highlighted in the evaluation performance, where the proposed method holds the pole for more than 50 steps in the average case contrary to the other evaluated baselines.

\begin{figure}[t]
    \centering
    \includegraphics[width=\linewidth]{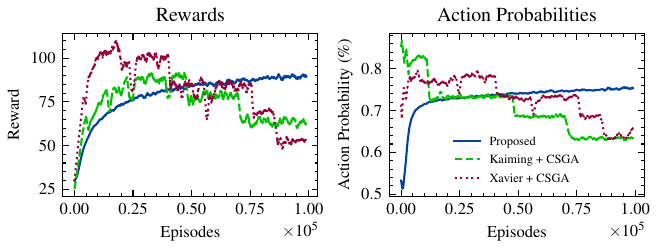}
    \caption{On the left, the figure depicts the obtained reward during training that is smoothed using a moving average filter with a window of 100. On the right, the action probabilities for each method are depicted using the same moving average setting.}
    \label{fig:performance}
\end{figure}

To highlight such different behavior, we report in Figure~\ref{fig:performance}, the average reward between different runs and the average action probability during training. For each case, we smooth the reported result using a moving average filter, setting the window value equal to 100. Similarly to the results reported in Table~\ref{tab:results}, we observe that during training, the two baselines are highly unstable, resulting in a poor local minimum and, as a result, significantly lower rewards. On the other hand, the proposed method allows for more consistent training, achieving significantly higher performance than the baselines.

This instability in training can also be observed in the action probability plot, where the baselines lead to increases at a high rate during the initial stage of training,  before subsequently gradually decreasing the corresponding probabilities. This indicates that baselines converge quickly to a bad local policy, with the clipping update term introducing difficulties in the training process. Such difficulties can be attributed to the fact that the clipping method zeros out synapses when they try to change sign, reducing the learning capacity of the model. This can also lead to vanishing gradient phenomena, which in turn can lead to bad local minima or even halt the training process~\cite{kirtas2022robust}. On the other hand, the proposed optimization method ensures that the parameters will remain non-negative without suppressing weights to zero, allowing gradients to flow through the network since the absolute value operator has a non-zero derivative both for positive and negative values. In this way, it provides a smooth training process and consistent results, as well as a more explainable representation for the actor model, allowing one to extract rationales of the agent due to the non-negative constraints that diminish canceling neurons and leading to part-based representations.

\begin{figure}[t]
    \centering
    \includegraphics[width=\linewidth]{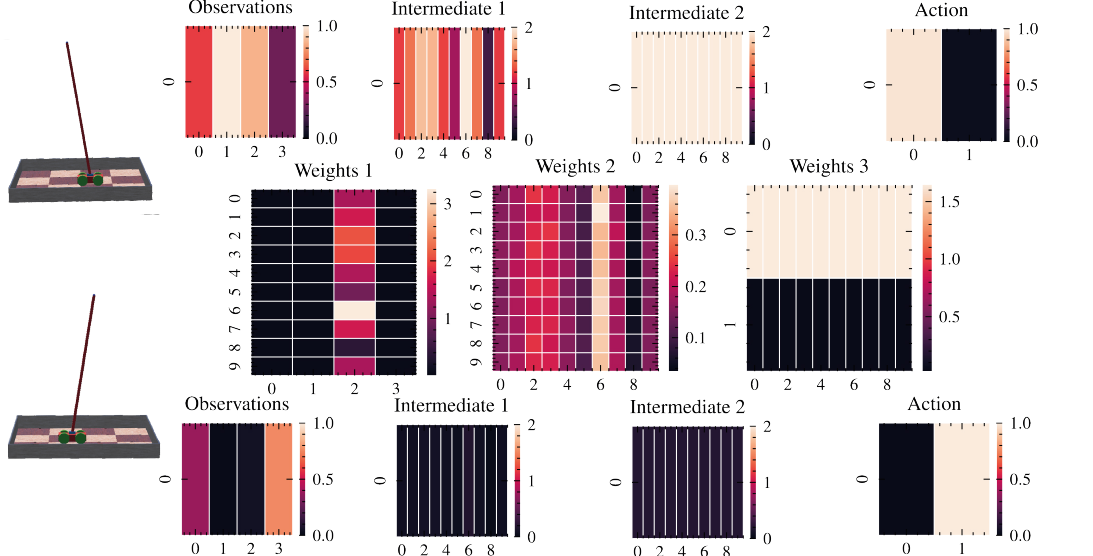}
    \caption{Part-based representation of the actor model. At the top and bottom rows, the input and responses of each layer are depicted. In the central row, the weights of the actor model are depicted. The biases are omitted for simplicity.}
    \label{fig:part_based}
\end{figure}


To highlight this, in Figure~\ref{fig:part_based} we provide an example of the part-based representation acquired using the proposed method. More specifically, two cases are presented: a) when the pole is dropped from the front of the cart (top row) and b) when the pole is falling from the rear side of the cart (bottom row). On the middle row, the weights of the model are depicted, omitting biases for simplicity. The observations are presented on the first plot from the left and are normalized between [0, 1], with columns on the plot denoting from 0 to 3: 0) cart position, 1) cart velocity, 2) pole angle and 3) endpoint velocity. The action made by the actor is represented on the right plot with the zeroth neuron response denoting the forward move of the agent (left as depicted in the figure) and the first neuron the backward move (right as depicted in the figure).

As depicted in Figure~\ref{fig:part_based}, applying part-based learning by using the proposed method allows one to extract visually meaningful representations. More precisely, when the pole is dropped from the positive side of the x-axis, meaning on the front side of the cart, and has large values for the pole angle, the agent moves the cart toward the same direction as the falling pole trying to keep it vertical, as depicted in the first row of the figure with the lighter heatmaps. On the other hand, when the pole has lower values on the pole angle (darker colors in the observations' heatmap are used to denote this), the agent moves the cart backward. Lighter heatmaps when multiplied with the final layer's weights (weights 3 in the figure) of the actor lead to firing the zeroth neuron of the layer that is translated on moving the cart forward. We can safely conclude that the agent has high confidence in its decision by comparing the intermediate representation of two cases after the second layer, since the two representations are significantly different. As depicted, in the first case (first row of the figure), the last layer zeroes out the response of the first neuron, maintaining the large values on the input of the zeroth neuron. On the contrary, in the second case, the input of the last layer already has values close to zero, with the first layer increasing the values of the input to fire the zeroth neuron. We have to mention that the action heatmap of the figure is after the softmax activation function, translating the response of the last layer to action probabilities.

We can also extract the rationale for the agent based on other observations. For example, the second observation (meaning the column denoted by 1 in the figure) refers to the cart velocity (normalized between range [0, 1]). This means that if the value of cart velocity is lower than 0.5, then the cart has a direction to the negative side of the x-axis (backward movement). Accordingly, if the cart velocity value is greater than 0.5, then the cart is moving forward, towards the positive side of the x-axis. As depicted in the observations' heatmap, the agent moves toward the opposite direction of the direction that the pole is falling. This is expected since the observations are extracted from an irrational agent, where the actor parameters have been randomly initialized. However, the trained agent using the proposed method outputs actions that moves the cart in the same direction as the falling pole to keep the pole vertical. This can also be observed in the heatmap of the weights of the first layer, where the third column has an inverse color with respect to the second column. Indeed, there is an inverse correlation between the cart position and the pole's angle.

\begin{figure}[htb!]
    \centering
    \includegraphics[width=0.8\linewidth]{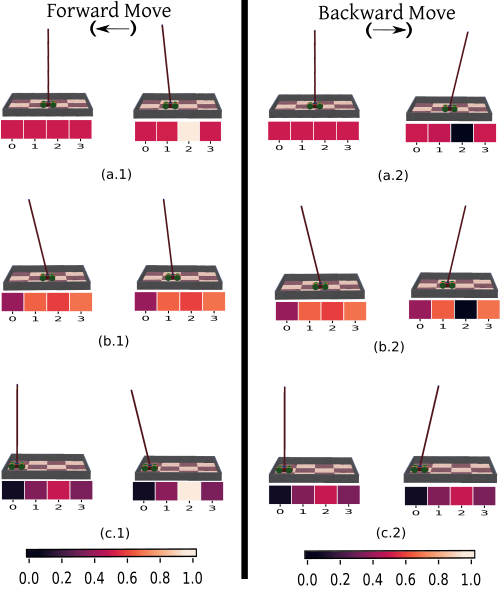}
    \caption{Optimized input of actor model to maximize the action probability of a given action using three different initialization. In the first column, the observations are optimized to maximize the forward action probability. In the second column, the observations are optimized to maximize the backward action probability. A different initialization of the observation vector is used for each row.}
    \label{fig:backward}
\end{figure}

We extend our analysis to obtain further insight regarding the response of the actor model, leveraging the advantages provided by the part-based representation of the model. To this end, we optimized the observation vector keeping the trained weights frozen in order to minimize the distance between the output of the model and a given action. The inputs are optimized for 5 epochs using SGD optimizer with the proposed sign-preserving update function applying mean squared error loss. We applied the proposed optimization method to ensure that the values of the observations will remain non-negative according to the actual normalized observations acquired from the environment. The observations obtained are presented in Figure~\ref{fig:backward}, where in the left column they are optimized to maximize the action probability of the forward action, and, respectively, on the right side they are optimized to maximize the action probability for the backward action. For each case, we report both the initial vector of observation and the one after the optimization process.

To evaluate which of the observations is more significant for each action, we initialize each observation to 0.5 and then optimize them to maximize the respective action probability, depicted in the (a) row. As expected, for both actions, the observation that changed significantly is the pole angle (in the third column). Indeed, when the observations are optimized in order to obtain a forward action from the actor model, the pole angle value converges to its maximum, which is value one. On the other hand, when the observations are optimized in order to obtain a backward action, the pole angle value converges to a value close to zero. In both cases, the other observations remain close to the initialized values, and it seems that they are not significantly affected by the optimization process. This is an expected behavior since the pole angle is the most significant indicator, revealing the direction in which the pole is falling, and as a result the one that significantly contributes to the prediction of the actor model.

We repeat the aforementioned experimental setup by applying different initializations on the observations vector and reporting the results at the (b) and (c) rows. More specifically, we randomly draw values from a Gaussian distribution with 0.5 mean and 0.2 standard deviation. As shown, except for the pole angle observation, which is significantly changed in most cases, the rest of the observations are not affected by the optimization process. Regarding the observation of the pole angle, when it has opposite direction with the given action, the obtained direction of the pole angle after optimization is inverted, such in case (b.2). In cases where the initialized observation vector has the same direction of the pole angle with the given action, as in cases of (b.1), the optimization process slightly changes the initial vector and the value of the pole angle. Finally, it is observed that when the initialized pole is vertical, such as in (a) and (c) cases, the optimization process leads to maximizing or minimizing the pole angle value according to the given action, resulting in the same direction with the action.    

\section{Conclusions}
\label{sec:conclusions}

In this study, we have introduced a novel training approach that focuses on non-negativity in deep RL using PPO. The proposed approach enables the extraction of part-based representations, which offers enhanced interpretability while following non-negative constraints associated with human cognition. To achieve this objective, the proposed method employs a non-negative initialization technique, followed by a modified sign-preserving training method. More specifically, we proposed employing an exponential distribution-based non-negative initialization method for the actor model and then using an appropriately modified sign-preserving alternative to Stochastic Gradient Ascent (SGA) for training the actor model in a non-negative manner. By adopting the proposed method, we can mitigate issues related to the reduction of the learning capacity of models and the vanishing gradients due to the use of clipping mechanisms involved in existing approaches. This helps mitigate issues such as vanishing gradients and enhances training stability. Consequently, the proposed pipeline enables more efficient training of inherently explainable models based on the non-negative part-based representation of the actor. To validate the effectiveness of the proposed method, we conducted experiments on the well-established Cartpole benchmark. The results demonstrate the effectiveness of the proposed method in achieving superior performance and showcasing the advantages of the proposed non-negative training methodology.

The promising results reported in this paper highlight several interesting future research directions. First, the proposed method can also be extended to handle value-based RL approaches, such as DQN~\cite{mnih2013playing}. Furthermore, extending the part-based representation learning to the actor model could also provide further insight into the training dynamics of the RL process, as well as allow for better adapting it to the task at hand, potentially leading to more robust algorithms. Finally, combining the proposed method with distillation approaches that can transfer knowledge from the intermediate layers of traditional DL models, for example~\cite{passalis2018learning}, could potentially allow for better guidance of the optimization process and learning more accurate policies.

\section*{Acknowledgments}
This work was supported by the European Union’s Horizon 2020 Research and Innovation Program (OpenDR) under Grant 871449. This publication reflects the authors’ views only. The European Commission is not responsible for any use that may be made of the information it contains.

\bibliographystyle{ieeetr}
\bibliography{bibliography}

\end{document}